\documentclass[conference]{IEEEtran}
\IEEEoverridecommandlockouts

\usepackage[utf8]{inputenc}
\usepackage[T1]{fontenc}
\usepackage{amsmath,amssymb,amsfonts}
\usepackage{graphicx}
\usepackage{booktabs}
\usepackage{multirow}
\usepackage{adjustbox}
\usepackage{hyperref}

\usepackage{xcolor}

\newcommand{\myParagraph}[1]{\textbf{#1.}\quad}

\hypersetup{
    colorlinks=true,
    linkcolor=blue,
    urlcolor=cyan,
    pdftitle={STREAM-VAE},
}

\ifCLASSOPTIONcompsoc
  \usepackage[nocompress]{cite}
\else
  \usepackage{cite}
\fi

\graphicspath{{.}{./figs/}}

\makeatletter
\def\ps@IEEEtitlepagestyle{%
  \def\@oddfoot{\mycopyrightnotice}%
  \def\@evenfoot{}%
}

\def\mycopyrightnotice{%
  {\footnotesize
  \begin{minipage}{\textwidth}
  \centering
  \textcopyright~2026 IEEE. Personal use of this material is permitted.
  Permission from IEEE must be obtained for all other uses, in any current
  or future media, including reprinting/republishing this material for
  advertising or promotional purposes, creating new collective works,
  for resale or redistribution to servers or lists, or reuse of any
  copyrighted component of this work in other works.

  \vspace{2pt}

  Published in the 2026 IEEE Intelligent Vehicles Symposium (IV),
  Detroit, MI, USA, pp. 444--451.
  DOI: \href{https://doi.org/10.1109/IV66570.2026.11623993}%
  {10.1109/IV66570.2026.11623993}
  \end{minipage}}%
}
\makeatother

\begin{document}

\title{STREAM-VAE: Dual-Path Routing for Slow and Fast Dynamics in Vehicle Telemetry Anomaly Detection}

\author{%
  Kadir\mbox{-}Kaan Özer\IEEEauthorrefmark{1}\IEEEauthorrefmark{2},
  Ren\'{e} Ebeling\IEEEauthorrefmark{1},
  Markus Enzweiler\IEEEauthorrefmark{2}%
  \thanks{%
    \IEEEauthorrefmark{1}Mercedes-Benz AG, Germany.\par
    \IEEEauthorrefmark{2}Institute for Intelligent Systems,
    Esslingen University of Applied Sciences, Germany.%
  }%
}

\maketitle

\begin{abstract}
Automotive telemetry data exhibits slow drifts and fast spikes, often within the same
sequence, making reliable anomaly detection challenging. Standard
reconstruction-based methods, including sequence variational autoencoders (VAEs), use a single latent
process and therefore mix heterogeneous time scales, which can smooth out spikes
or inflate variances and weaken anomaly separation.

In this paper, we present STREAM-VAE, a variational autoencoder for anomaly detection in automotive telemetry time-series data. Our model uses a dual-path encoder to separate slow drift and fast spike
signal dynamics, and a decoder that represents transient deviations separately from the
normal operating pattern. STREAM-VAE is designed for deployment,
producing stable anomaly scores across operating modes for both in-vehicle
monitors and backend fleet analytics.

Experiments on an automotive telemetry dataset and the public SMD benchmark show that
explicitly separating drift and spike dynamics improves robustness compared to
strong forecasting, attention, graph, and VAE baselines.
\end{abstract}


\section{Introduction}
Modern intelligent vehicles stream high-rate telemetry signals from powertrain,
chassis, ECUs, and body controllers. Detecting abnormal patterns in these
signals is important for identifying emerging faults early and ensuring
reliable vehicle operation, both in on-board monitoring and in backend fleet
analytics. Yet several properties make this task difficult. The data are high
dimensional and contain strong cross-sensor couplings. Vehicles operate in
nonstationary modes that depend on traffic, environment, and driver behavior.
Abnormal behavior may appear as brief transients caused by driver input or
electrical disturbances, or as slow-evolving drifts driven by load or
temperature changes.

\begin{figure}[t]
  \centering
   \includegraphics[width=1.\linewidth]{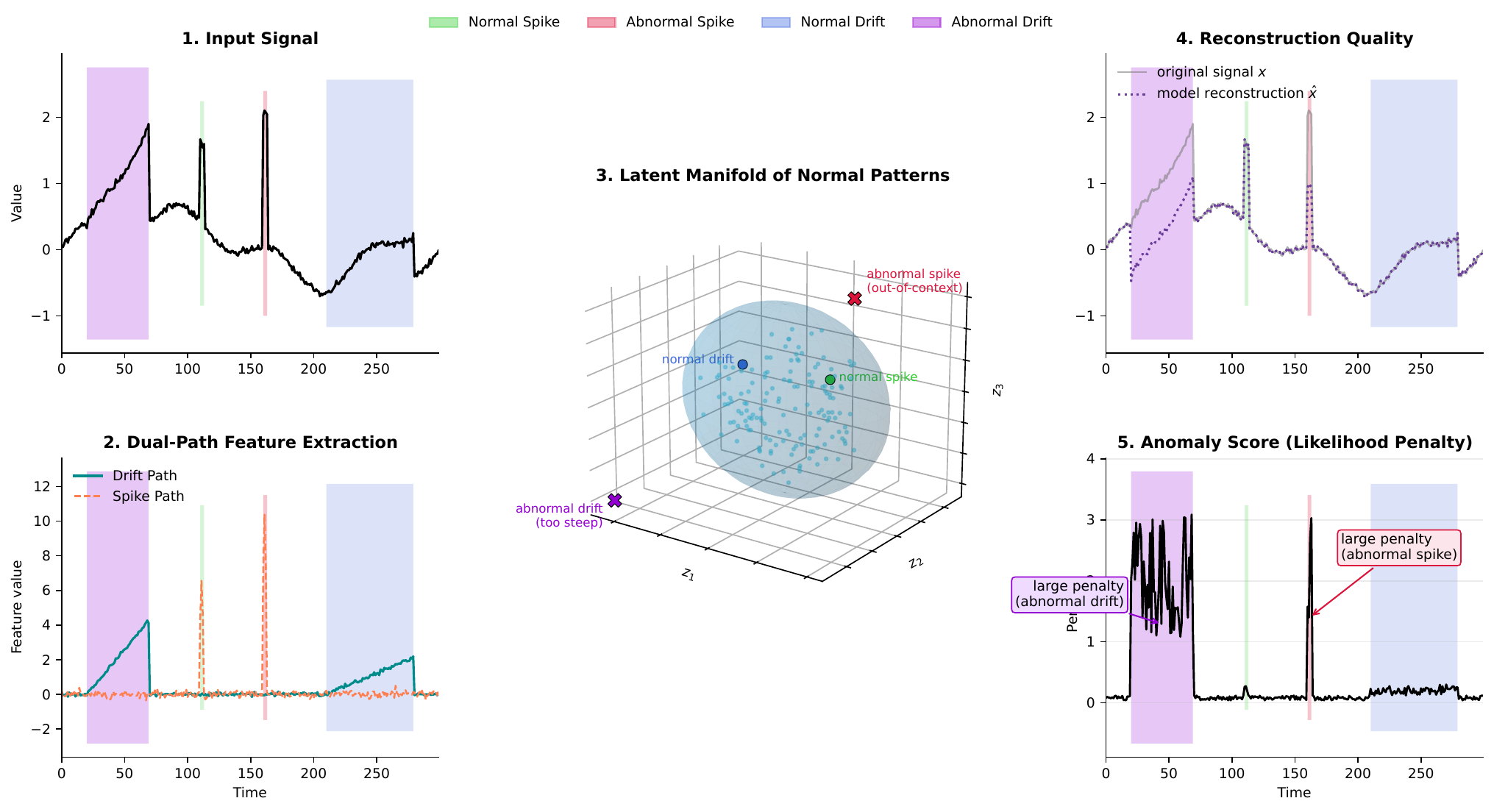}
   \caption{Illustration of the core detection challenge: fast spikes and slow drifts distort different aspects of the signal. STREAM-VAE maps clean segments to a compact latent manifold, while anomalous patterns should fall outside this region.} 
   \label{fig:signals}   
\end{figure}

Variational autoencoders (VAEs) are widely used in reconstruction-based anomaly
detection, where a model is trained on normal data and high reconstruction
error indicates abnormal behavior. They also provide a probabilistic notion of
expected behavior through a learned likelihood~\cite{kingma2013vae}. In practice, time-series VAEs encode each temporal segment into a single latent trajectory, requiring the model to capture both high-frequency variations and low-frequency drifts within a shared representation. This often leads to over-smoothed spikes or inflated variances that weaken the separation between nominal and anomalous scores (see Fig.~\ref{fig:signals}). For deployment in intelligent vehicles, this separation matters both for lightweight in-vehicle monitors and for high-volume backend fleet analytics, where stable thresholds and consistent behavior across operating modes are critical.

Prior work has addressed parts of this challenge. Smoothing regularizers stabilize reconstructions but can attenuate short events~\cite{li2021sisvae}. Attention-based encoders capture long-range structure~\cite{zhao2022dualstage,pereira2018vsvae,bahuleyan2018}, although they do not enforce any separation between fast and slow components and therefore mix heterogeneous time scales in a single latent representation. Automotive-oriented designs show how to prevent attention from bypassing the latent bottleneck~\cite{correia2023MAVAE}, but they do not provide a mechanism for handling transient spikes separately from slow drifts. Posterior collapse remains a risk with expressive decoders, motivating feedback-based KL control~\cite{shao2020controlvae}. Outside the VAE family, strong baselines such as TFT-Residual~\cite{lim2021tft}, Anomaly Transformer ~\cite{xu2022anomalytransformertimeseries}, and GDN~\cite{deng2021gdn} perform well on specific data regimes. However, they typically rely on residual thresholds or per-series tuning, and they do not directly address the need for consistent modeling of mixed time-scale behavior. In summary, existing methods lack an explicit mechanism to separate fast and slow dynamics and to explain transient spikes without widening the nominal tail.

\begin{figure*}[!t]
  \centering
  \includegraphics[width=0.9\textwidth]{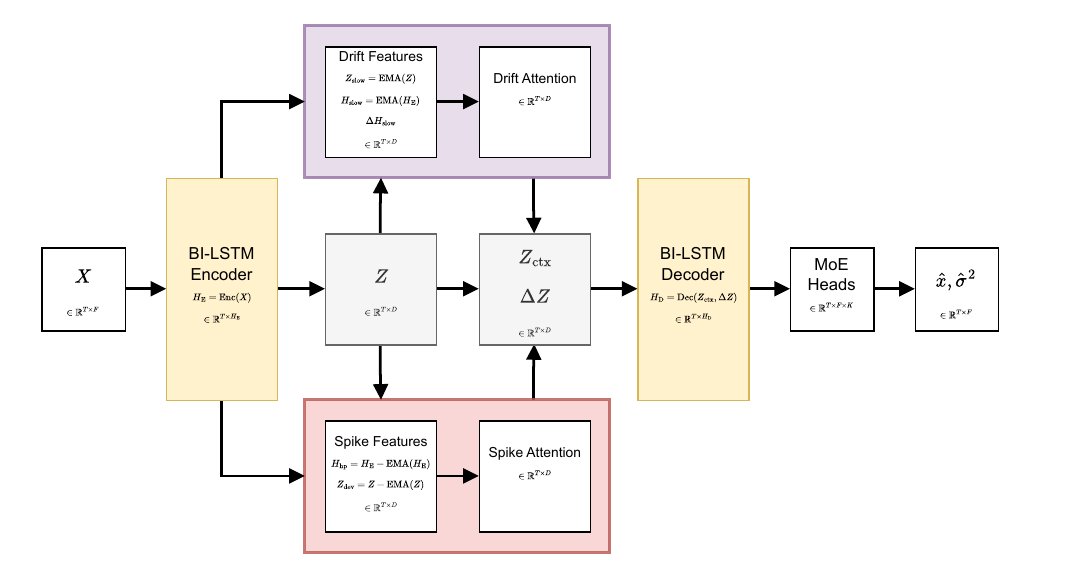}
  \caption{Overview of the STREAM-VAE architecture with dual drift (top, purple) and spike (bottom, red) paths. Each block is annotated with the shape of its output tensor. The BI-LSTM Encoder (yellow) maps the input sequence $X \in \mathbb{R}^{T \times F}$ to encoder states $H_E \in \mathbb{R}^{T \times H_{\text{E}}}$ and latent states $Z \in \mathbb{R}^{T \times D}$. Slow EMA-based features define the drift path, while high-pass residual features define the spike path; both produce attention outputs in $\mathbb{R}^{T \times D}$. A gated fusion yields the latent context $Z_{\text{ctx}} \in \mathbb{R}^{T \times D}$, and the first difference $\Delta Z \in \mathbb{R}^{T \times D}$ carries transient information. The BI-LSTM Decoder maps these to hidden states $H_D \in \mathbb{R}^{T \times H_{\text{D}}}$, which drive per-feature mixture-of-experts (MoE) heads that output Gaussian reconstruction parameters $(\hat{x}, \hat{\sigma}^2) \in \mathbb{R}^{T \times F}$ used for anomaly scoring.}
  \label{fig:streamvae}
\end{figure*}

Our main contribution in this paper is STREAM-VAE, which stands for
\textbf{S}pike \textbf{T}rend \textbf{R}outing with \textbf{E}vent Residual
\textbf{A}ttention and \textbf{M}ixture of Experts \textbf{VAE}. The model
introduces a dual-path architecture that separates slow drift and fast spike
feature dynamics in the encoder through two individual attention paths~\cite{vaswani2023attentionneed}. The decoder combines a per-feature
mixture of experts (MoE) with an event-residual block that pairs a residual connection~\cite{he2016deep}
with soft-thresholding~\cite{donoho1995denoising} so that transient
deviations can be represented without widening the nominal likelihood tail.
These design choices directly address the two limitations outlined above: the
absence of explicit slow-fast separation and the difficulty of representing
brief spikes without inflating nominal tails. The architecture is intended to
support both in-vehicle monitoring, where models must remain compact and
predictable, and backend fleet-level analysis, where stable tail behavior
enables consistent thresholding across vehicles. See
Fig.~\ref{fig:streamvae} for an overview.

\section{Related Work}
Work on time-series anomaly detection in vehicle telemetry spans three lines that motivate our choices, especially in the context of automotive sensor streams with mixed time scales and frequent mode changes.

First, classical and forecasting-residual approaches treat anomalies as departures from a predictive signal. Isolation Forest isolates points via random partitions and remains competitive when feature scales are well behaved~\cite{liu2008isolation}. Sequence forecasters such as the Temporal Fusion Transformer with a residual scoring head (``TFT-Residual'') fit rich conditional dynamics and flag large residuals~\cite{lim2021tft}, but residuals  can overreact to mode switches (e.g., gear changes, drive-cycle transitions between urban and highway segments, or sudden load steps), which are common in vehicle telemetry unless operating mode changes are modeled explicitly. 

A second line of work emphasizes self-attention and structure-aware discrepancy. Anomaly Transformer scores each timestamp by how atypical its associations are relative to learned patterns, yielding sharp localization on periodic or quasi-periodic series~\cite{xu2022anomalytransformertimeseries}. Graph-based methods such as GDN encode inter-sensor relations and measure deviations from graph-conditioned expectations, which is effective when the sensor topology is stable~\cite{deng2021gdn}, although automotive subsystems often exhibit context-dependent couplings.

A third complementary line of research builds probabilistic generative models that return calibrated likelihoods. OmniAnomaly uses a stochastic recurrent VAE with flow-based posteriors~\cite{su2019omni,rezende2015flows}. Li et al.~\cite{li2021sisvae} add smoothing to stabilize reconstructions. MA-VAE integrates attention while protecting the latent bottleneck~\cite{correia2023MAVAE}. Models utilizing variational self-attention~\cite{pereira2018vsvae}, Wasserstein distance–based
latent anomaly scores~\cite{pereira2019wvae}, and sparse/structured VAEs such as VASP~\cite{he2021vasp} further improve optimization and latent selectivity. These models provide calibrated scores, but a single latent process is typically forced to explain both driver-induced spikes and slow environmental drifts, which is problematic for in-vehicle monitoring where the two have distinct diagnostic meaning. For brevity, we will refer to~\cite{pereira2018vsvae} as VS-VAE,~\cite{pereira2019wvae} as W-VAE (Wasserstein\mbox{-}similarity scoring), and~\cite{li2021sisvae} as SIS-VAE.

Across these model families, progress has improved context modeling, graph structure, expressivity, and likelihood calibration. Yet two frictions remain when deploying to intelligent vehicles: (i) a single latent path is often required to explain both spikes and drifts, and (ii) many systems depend on per-series thresholds, which complicates fleet-wide calibration. 

STREAM-VAE maintains a probabilistic likelihood but separates fast and slow content in a way that aligns with both actuator-level dynamics and operating mode shifts. Additionally, it decodes with a per-feature mixture of experts plus a soft-thresholded event path so that transients do not broaden the nominal tail.

\section{Model Architecture}
\label{sec:model}

Modern vehicle telemetry exhibits two types of changes that matter for detection:
very short, high-amplitude \emph{spikes} (e.g., pedal jabs or brief current
surges) and slower \emph{drifts} (e.g., load or temperature trends and operating mode 
switches). If a single latent stream is forced to explain both, spikes are
often over-smoothed or the decoder inflates variance to cover drifts, in both
cases narrowing the gap between nominal and anomalous scores. STREAM-VAE
addresses this by (i) routing fast and slow content through two encoder paths
and (ii) decoding with a per-feature mixture of experts that absorbs benign
mode changes, plus a gated event residual that explains sparse transients
\emph{without} broadening the nominal likelihood tail. Fig.~\ref{fig:streamvae}
shows the full architecture; we now describe each component following the
left-to-right flow of the diagram.

\myParagraph{Input $X$}
We process standardized windows of length $T$ with $F$ features. The encoder produces a per-timestep latent mean and variance logit in $D$ dimensions. Logits are mapped to positive values by softplus and clipped for numerical stability. To keep the latent representation informative
even when the decoder is expressive, we add a small learnable linear projection
of the raw input into the posterior heads. This projection is initialized to
zero and gated by a sigmoid so that it influences the posterior only if it
proves useful during training.

\myParagraph{BI-LSTM Encoder}
A light two-layer BI-LSTM trunk~\cite{GRAVES2005602} maps the input window to contextual
representations for each timestep. These features feed the posterior heads,
which produce the latent sequence $Z$. The same encoder features also feed the
two attention branches described below.

\myParagraph{Latent $Z$}
From the latent sequence $Z$ we form the first difference $\Delta Z$ and a
smoothed sequence $\mathrm{EMA}(Z)$. Here $\mathrm{EMA}(\cdot)$ denotes an exponential moving average applied along the time dimension with a learnable smoothing factor. The difference $\Delta Z$ later drives the event residual in the decoder, and $\mathrm{EMA}(Z)$ supplies the value stream
for the slow branch so that gradual trends are modeled directly.

\myParagraph{Drift Features}
Let $H_{\text{E}} = \mathrm{Enc}(X)$ denote the encoder features and $Z$ the corresponding
latent sequence. To emphasize slow evolution, we apply a (separately learned) EMA to obtain a
slowly varying baseline $H_{\text{slow}} = \mathrm{EMA}(H_{\text{E}})$ and then take a
first difference along time to obtain $\Delta H_{\mathrm{slow}}$. Linear
projections of $\Delta H_{\mathrm{slow}}$ form the queries and keys for the
drift attention branch, and the values come from the smoothed latent sequence $Z_{\mathrm{slow}} = \mathrm{EMA}(Z)$ so that this branch follows gradual trends rather than noise.

\myParagraph{Spike Features}
To expose brief and localized deviations, we compute a high-pass
residual
$H_{\text{hp}} = H_{\text{E}} - \mathrm{EMA}(H_{\text{E}})$
on the encoder features using a learnable EMA baseline. Linear projections of
$H_{\mathrm{hp}}$ provide the queries and keys for the spike attention branch,
while the values use the corresponding deviation in latent space
$Z_{\mathrm{dev}} = Z - \mathrm{EMA}(Z)$.

\myParagraph{Drift Attention}
Multi-head attention uses queries, keys, and values from the Drift Features. This produces a $D$-dimensional representation that focuses on slow and
persistent context. Queries and keys are $\ell_2$-normalized so that attention matching is scale-invariant.
Grouped or multi-query attention (GQA) is used to reduce key–value memory cost without changing model behavior~\cite{ainslie2023gqa}.  

\begin{figure*}[t]
  \centering
  \includegraphics[width=\linewidth]{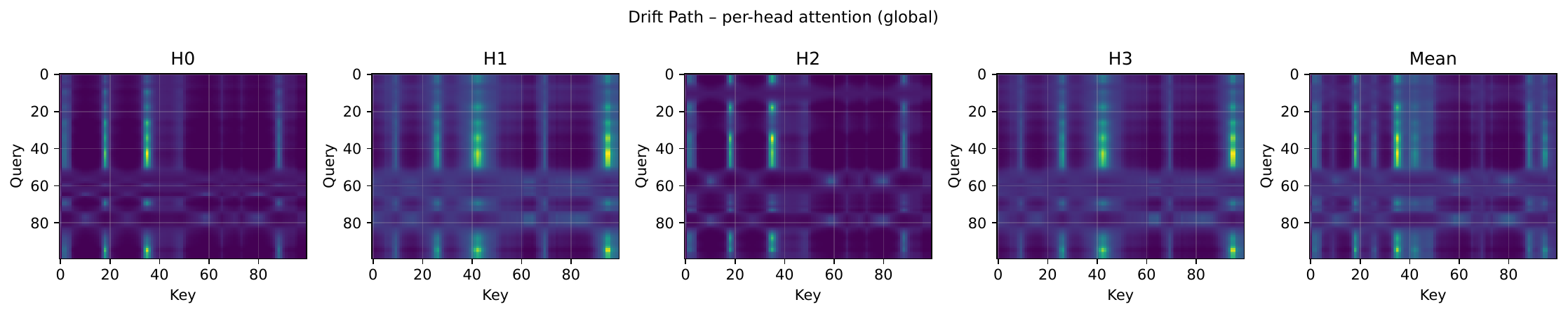}
  \includegraphics[width=\linewidth]{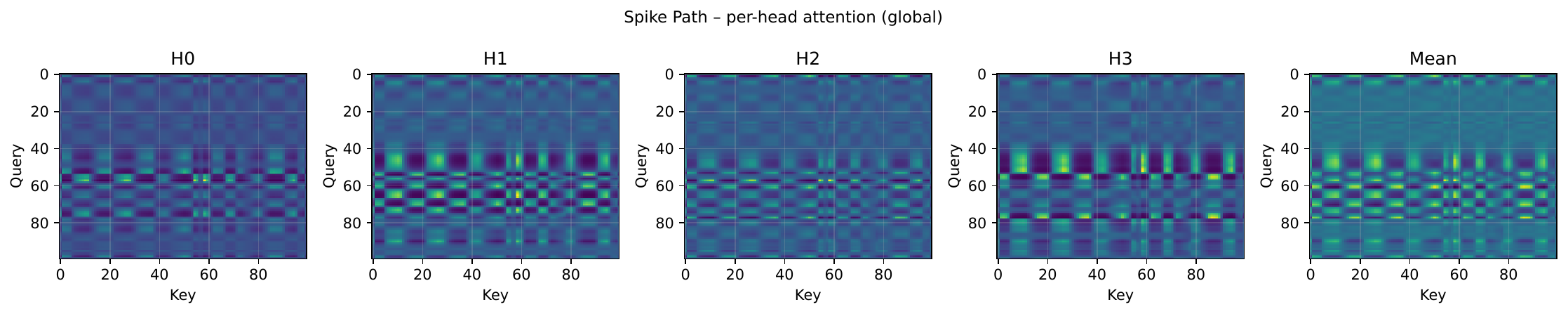}
\caption{
Dual-path global attention (per head). 
\textbf{Drift path (top).}
Multiple heads show bright vertical bands at consistent key indices, so many
queries align to the same columns. This produces a few global reference
anchors and a low-rank, anchor-based alignment that reflects slow and
persistent context. 
\textbf{Spike path (bottom).}
Heads display dense, oscillatory checkerboard patterns and narrow 
query-centric stripes, which indicate short-lag, high-frequency coupling
typical of localized transients. The mean maps remain structured in both
rows, which shows that heads specialize consistently and that the two paths
split cleanly: The drift path captures global, low-frequency drift, and the spike path captures localized, high-frequency spikes.
}
  
  \label{fig:global_heads}
\end{figure*}

\myParagraph{Spike Attention}
A parallel branch uses queries, keys, and values from the Spike Features to target brief and localized transients. Fig.~\ref{fig:global_heads} illustrates
the distinct attention patterns learned by the drift and spike attention pathways.

\myParagraph{Decoder Input ($Z_{\text{ctx}}$, $\Delta Z$)}
The decoder uses two latent signals. A sigmoid gate blends the drift and spike
outputs at each timestep, and a light residual block refines the mixture to
form the final encoder context $Z_{\text{ctx}}$, which is fed to the BI-LSTM
decoder. In parallel, we take the first difference of the latent sequence to
obtain $\Delta Z$, which is used for the event-residual.

\myParagraph{BI-LSTM Decoder}
The decoder is a two-layer BI-LSTM that maps $Z_{\text{ctx}}$ to hidden states used to
parameterize the reconstruction mean and variance. Brief transients are modeled
additively in the mean through an event residual driven by the first
difference $\Delta Z$. A linear map produces a per-feature residual, which is
soft-thresholded with a per-feature shrinkage parameter so that small activations are suppressed. A global gate and per-feature gains
control the magnitude of this residual, and an RMS-based scaling keeps it
commensurate with the MoE base mean. The variance head is a single linear layer
followed by a softplus and clipping, and is shared across experts so that
variance does not absorb transient spikes.

\myParagraph{MoE Heads}
For each timestep and feature, the decoder uses its current hidden state to
produce non-negative softmax weights over the $K$ experts. From that same
hidden state, it also generates $K$ expert means (via a low-rank
factorization), and forms a base mean by mixing them with the softmax weights.
This routing lets the model follow different nominal operating modes by
shifting weight across experts instead of broadening the likelihood tails,
consistent with classic mixture-of-experts ideas~\cite{jacobs1991adaptive,
jordan1994hmoe}. In practice, only a few experts (e.g., $K{=}2$--$4$) are
sufficient.

\myParagraph{Output ($\hat{x}$, $\hat{\sigma}^2$)}
The final decoder mean $\hat{x}$ is the sum of the base mean from the MoE heads
and the gated, soft-thresholded event residual. The variance $\hat{\sigma}^2$ is produced by a
shared variance head. The anomaly score for each window is the Gaussian negative log-likelihood, where larger values indicate more anomalous behavior. Labels are aligned to the end of each window so that scores and
labels refer to the same temporal point.

\myParagraph{Training Objective and KL Control}
Training minimizes a Gaussian reconstruction term, a KL term with a feedback-controlled coefficient $\beta$ to keep the latent informative, and two light regularizers: an $\ell_1$ penalty on the event residual to encourage sparsity and a weak entropy target on the MoE gates to prevent expert collapse without forcing uniformity:
\begin{equation}
\label{eq:loss}
\begin{aligned}
\mathcal{L}
=\; & -\log p_\theta(X\mid Z)
\;+\; \beta\, D_{\mathrm{KL}}\!\big(q_\phi(Z\mid X)\,\|\,\mathcal{N}(0,I)\big) \\
& +\; \lambda\,\|r\|_1 \;+\; \eta\,\big(H^\star - H\big)_{+}.
\end{aligned}
\end{equation}
Here $H$ is the average entropy of the MoE gates and $H^\star$ is a mid-level
target equal to half of $\log K$. We use $(x)_{+} = \max(x,0)$ to denote the
positive part. The coefficient $\beta$ is updated online with
a proportional controller and EMA smoothing, following controllable VAE
methods~\cite{shao2020controlvae}. Optimization uses Adam with gradient-norm
clipping~\cite{kingma2015adam}.

\myParagraph{Scoring and Thresholding}
Thresholds are calibrated \emph{once} per time series on \emph{normal} training windows via Peaks-Over-Threshold (POT)~\cite{coles2001extreme}, which models exceedances
with a Generalized Pareto Distribution (GPD). To make this reliable during model selection (see Sec.~\ref{sec:experimental_setup}), we penalize heavy nominal tails using a normalized upper-quantile width and a GPD-shape penalty evaluated on validation normals, using robust scale
estimators to avoid outlier bias~\cite{rousseeuw1993alternatives}.

\myParagraph{Implementation Notes}
EMA memory factors for the drift and spike baselines are learned and initialized empirically at $0.9$, a conventional choice that provides a smooth yet responsive starting point for separating slow and fast components, e.g.,~\cite{kingma2015adam}. All smoothing and differencing operations are applied to the specific signal used in each branch, following the routing shown in Fig.~\ref{fig:streamvae}. Attention heads are split evenly across drift and spike branches and can optionally use GQA~\cite{ainslie2023gqa}. MoE expert means are implemented with a
low-rank parameterization, and the event-residual thresholds are
softplus-parameterized and initialized to small values. The shared variance head uses clipping for numerical stability. A feedback controller maintains the KL term near its target value and removes the
need for manual schedule tuning~\cite{shao2020controlvae}.

\section{Experimental Setup}
\label{sec:experimental_setup}


We evaluate STREAM\mbox{-}VAE and strong baselines on a proprietary automotive
telemetry dataset and the public Server Machine Dataset (SMD)~\cite{su2019omni}. Each dataset provides a train/test split per entity, and anomalies are labeled.

\myParagraph{Baselines}
We compare STREAM\mbox{-}VAE against a diverse set of strong baselines that are 
representative of current practice in time-series anomaly detection. The set includes
sparse and structured VAEs (VASP~\cite{he2021vasp}, VS-VAE~\cite{pereira2018vsvae},
W-VAE~\cite{pereira2019wvae}, SIS-VAE~\cite{li2021sisvae}, MA-VAE~\cite{correia2023MAVAE}),
a stochastic recurrent VAE with flow-based posteriors (OmniAnomaly~\cite{su2019omni}),
a graph-based detector (GDN~\cite{deng2021gdn}), an attention-based forecaster with 
residual scoring (TFT-Residual~\cite{lim2021tft}), a structure-aware attention model 
(Anomaly Transformer~\cite{xu2022anomalytransformertimeseries}), and a classical tree 
ensemble (Isolation Forest~\cite{liu2008isolation}).

\myParagraph{Automotive Dataset}
Our automotive dataset consists of clean in-vehicle test-drive telemetry 
augmented with expert-designed synthetic anomalies that match field distributions 
and remain physically consistent. The dataset has been acquired in different locations 
and scenarios from a test vehicle fleet over a period of several days.  A controlled injection simulator introduces short, contiguous perturbations affecting one or more correlated features within their valid operating ranges. Each anomaly follows one of six canonical fault types 
(\emph{spikes}, \emph{drifts}, \emph{level shifts}, \emph{variance jumps}, 
\emph{flatlines}, \emph{correlation breaks}), with magnitudes and durations sampled 
from type-specific calibrated ranges. During training, 30\% of the nominal data are 
held out for validation. Corpus- and feature-level summaries are given in Table~\ref{tab:dataset_feature_stats}. \looseness=-1

\myParagraph{SMD (Server Machine Dataset)}
SMD is a widely used benchmark in general-purpose time-series anomaly detection 
but is not commonly used in automotive research. We include it to evaluate 
generalization outside the vehicle domain and to compare STREAM-VAE against 
established baselines under a standardized, publicly reproducible protocol. 
SMD contains 28 entities (server nodes, referred to as \textit{machines}) with multivariate metrics and labeled 
anomalies. Although the domain differs, it shares structural challenges with 
vehicle telemetry: mixed time scales, multi-sensor coupling, and infrequent transients.
For SMD we report macro averages across all entities.

\begin{table}[!t]
\centering
\caption{Dataset and feature statistics for the automotive dataset}
\label{tab:dataset_feature_stats}
\begin{tabular}{lccc}
\toprule
\multicolumn{4}{l}{\textbf{Dataset Summary}} \\
\midrule
\multicolumn{3}{l}{Total Records}                      & 40,000 \\
\multicolumn{3}{l}{Features (see below)}               & 8 \\
\multicolumn{3}{l}{Normal Instances}                   & 36,426 \\
\multicolumn{3}{l}{Anomalous Instances}                & 3,574 \\
\multicolumn{3}{l}{Anomaly Rate (\%)}                  & 8.94 \\
\multicolumn{3}{l}{Completeness (\%)}                  & 100.00 \\
\multicolumn{3}{l}{Outlier Percentage (\%)}            & 14.46 \\
\multicolumn{3}{l}{Range Coverage (\%)}                & 100.00 \\
\addlinespace
\midrule
\multicolumn{4}{l}{\textbf{Feature-Level Statistics}} \\
\midrule
\textbf{Feature} & \textbf{Mean} & \textbf{Std} & \textbf{Skewness} \\
\midrule
Brake Torque                          & 189.530  & 658.052 & 5.13 \\
Battery Current                       & -1.774   & 20.977  & 2.25 \\
Accelerator Pedal Position (Raw)      & 11.366   & 22.548  & 6.44 \\
Accelerator Pedal Position (OBD)      & 13.338   & 12.631  & 2.86 \\
Steering Wheel Angular Speed          & 0.034    & 37.294  & -0.54 \\
Vehicle Speed                         & 37.247   & 49.637  & 2.79 \\
Wheel RPM (Front Left)                & 286.966  & 359.348 & 1.41 \\
Wheel RPM (Rear Left)                 & 288.820  & 366.703 & 1.44 \\
\bottomrule
\end{tabular}
\end{table}


\myParagraph{Preprocessing}
STREAM-VAE processes standardized windows of length $T{=}100$ with $F$ features 
($F{=}8$ for the automotive dataset; SMD varies per entity). The encoder outputs 
per-timestep latent means and variance logits in $D{=}64$ dimensions; these values 
define the default configuration across all experiments unless otherwise noted.

All methods share identical segmentation and normalization. Each feature is 
z-scored using statistics from non-anomalous (nominal) training data only. Training uses 90\% 
window overlap (stride $10$), and evaluation uses 100\% 
window overlap (stride $1$). Labels are assigned 
to the window end to align scores and labels.


\myParagraph{Thresholding}
Thresholds are calibrated \emph{once} per entity using Peaks-Over-Threshold (POT) 
on nominal train scores and then held fixed. For the automotive dataset we use 
$(q,p){=}(10^{-3},10^{-4})$. For SMD we use group-specific $q$ values 
(machine\mbox{-}1: $10^{-3}$; machine\mbox{-}2: $2.5{\times}10^{-3}$; 
machine\mbox{-}3: $5{\times}10^{-3}$; all with $p{=}10^{-4}$).  
If exceedances are too sparse, we fall back to the empirical $(1{-}\alpha)$ 
upper quantile without labels~\cite{coles2001extreme,siffer2017anomaly}.

\myParagraph{Scoring}
All models are mapped to a scalar anomaly score per window to make them comparable. 
VAE-style models that output $(\boldsymbol{\mu},\boldsymbol{\sigma}^2)$ use the Gaussian 
negative log-likelihood; all others use mean squared reconstruction or forecast error, 
oriented so that larger scores indicate more anomalous behavior.


\myParagraph{Evaluation Metrics}
We evaluate all models using four standard anomaly-detection metrics. 
AUC-PR and AUC-ROC provide threshold-free assessments of ranking quality. 
Point-Adjusted F1 (PA-F1) is computed at the fixed POT threshold and treats any 
detection within an anomaly window as a correct hit~\cite{xu2018kpi}. 
We also report Oracle PA-F1, which selects the best achievable PA-F1 over all 
possible thresholds and serves as a diagnostic upper bound~\cite{su2019omni}.


\myParagraph{Hyperparameter Optimization}
For hyperparameter optimization, we allocate 20 Optuna~\cite{akiba2019optuna} trials (Tree-structured Parzen Estimator + MedianPruner) per model.  
The selected configuration is retrained with early stopping (up to 30 epochs; batch size 50), and results are reported as mean$\pm$standard deviation over 5 seeds. To align model selection with stable POT calibration, we minimize a label-free objective on validation normals:
\begin{equation}
\label{eq:tail_objective}
J = \tilde L_{\text{val}} + \lambda\,\mathrm{NUQ}_q + \gamma\,\xi_{+},
\end{equation}
where $\tilde L_{\text{val}}$ is the per-dimension normalized validation loss, 
$\mathrm{NUQ}_q$ is a robust upper-quantile width ($q{=}0.995$), and $\xi_{+}$ 
is the positive part of the GPD shape parameter estimated from exceedances.  Minimizing $J$ discourages heavy nominal score tails that destabilize POT and removes the need for manual KL tuning. We fix $(\lambda,\gamma){=}(0.01,0.05)$ for all datasets. For SMD we use a two-phase protocol: hyperparameter search on {machine-1-1}, {machine-2-1}, {machine-3-1}, then retrain 
and evaluate on all 28 entities with per-entity thresholds as above.

\section{Results}


\begin{table*}[t]
\centering
\caption{Performance of the proposed STREAM-VAE model and baselines on the automotive and SMD datasets. Mean $\pm$ standard deviation over 5 seeds. \textbf{Bold} indicates the best-performing method for each dataset.}

\label{tab:combined_oracle_auc}
\resizebox{\textwidth}{!}{%
\begin{tabular}{@{}lcccccccc@{}}
\toprule
\multirow{2}{*}{\textbf{Model}} 
& \multicolumn{4}{c}{\textbf{Automotive}}
& \multicolumn{4}{c}{\textbf{SMD}} \\ \cmidrule(rl){2-5} \cmidrule(l){6-9}
& \textbf{Oracle PA-F1} & \textbf{PA-F1} 
& \textbf{AUC-PR} & \textbf{AUC-ROC}
& \textbf{Oracle PA-F1} & \textbf{PA-F1} 
& \textbf{AUC-PR} & \textbf{AUC-ROC} \\
\midrule
Isolation Forest~\cite{liu2008isolation} 
& 0.568 $\pm$ 0.008 & 0.512 $\pm$ 0.021 & 0.133 $\pm$ 0.003 & 0.610 $\pm$ 0.002
& 0.861 $\pm$ 0.143 & \textbf{0.552 $\pm$ 0.339} & 0.280 $\pm$ 0.220 & 0.764 $\pm$ 0.121 \\

VASP~\cite{he2021vasp}               
& 0.641 $\pm$ 0.007 & 0.445 $\pm$ 0.000 & 0.120 $\pm$ 0.001 & 0.599 $\pm$ 0.004
& 0.837 $\pm$ 0.180 & 0.367 $\pm$ 0.348 & 0.313 $\pm$ 0.237 & 0.737 $\pm$ 0.133 \\

Anomaly Transformer~\cite{xu2022anomalytransformertimeseries} 
& 0.698 $\pm$ 0.022 & 0.380 $\pm$ 0.100 & 0.132 $\pm$ 0.014 & 0.602 $\pm$ 0.024
& 0.871 $\pm$ 0.137 & 0.399 $\pm$ 0.319 & \textbf{0.462 $\pm$ 0.272} & \textbf{0.881 $\pm$ 0.107} \\

W-VAE~\cite{pereira2019wvae}          
& 0.701 $\pm$ 0.017 & 0.386 $\pm$ 0.091 & 0.197 $\pm$ 0.008 & 0.633 $\pm$ 0.012
& 0.890 $\pm$ 0.119 & 0.427 $\pm$ 0.362 & 0.399 $\pm$ 0.263 & 0.800 $\pm$ 0.132 \\

VS-VAE~\cite{pereira2018vsvae}        
& 0.735 $\pm$ 0.016 & 0.466 $\pm$ 0.021 & 0.258 $\pm$ 0.024 & 0.704 $\pm$ 0.012
& 0.845 $\pm$ 0.149 & 0.471 $\pm$ 0.364 & 0.336 $\pm$ 0.268 & 0.773 $\pm$ 0.153 \\

OmniAnomaly~\cite{su2019omni}         
& 0.756 $\pm$ 0.009 & 0.663 $\pm$ 0.027 & 0.346 $\pm$ 0.008 & 0.686 $\pm$ 0.010
& 0.843 $\pm$ 0.193 & 0.400 $\pm$ 0.375 & 0.332 $\pm$ 0.270 & 0.744 $\pm$ 0.173 \\

MA-VAE~\cite{correia2023MAVAE}         
& 0.808 $\pm$ 0.037 & 0.723 $\pm$ 0.097 & 0.451 $\pm$ 0.027 & 0.743 $\pm$ 0.015
& 0.810 $\pm$ 0.176 & 0.423 $\pm$ 0.336 & 0.338 $\pm$ 0.216 & 0.790 $\pm$ 0.131 \\

SIS-VAE~\cite{li2021sisvae}            
& 0.824 $\pm$ 0.014 & 0.701 $\pm$ 0.047 & 0.380 $\pm$ 0.030 & 0.753 $\pm$ 0.017
& 0.717 $\pm$ 0.277 & 0.316 $\pm$ 0.310 & 0.226 $\pm$ 0.212 & 0.658 $\pm$ 0.170 \\

GDN~\cite{deng2021gdn}                
& 0.825 $\pm$ 0.009 & 0.760 $\pm$ 0.007 & 0.498 $\pm$ 0.011 & 0.744 $\pm$ 0.003
& 0.930 $\pm$ 0.088 & 0.496 $\pm$ 0.359 & 0.451 $\pm$ 0.263 & 0.833 $\pm$ 0.130 \\

TFT-Residual~\cite{lim2021tft}        
& 0.830 $\pm$ 0.004 & 0.781 $\pm$ 0.022 & 0.479 $\pm$ 0.019 & 0.750 $\pm$ 0.014
& 0.906 $\pm$ 0.114 & 0.442 $\pm$ 0.356 & 0.438 $\pm$ 0.253 & 0.831 $\pm$ 0.111 \\
\midrule
STREAM-VAE (ours)          
& \textbf{0.857 $\pm$ 0.024} & \textbf{0.794 $\pm$ 0.026} 
& \textbf{0.532 $\pm$ 0.030} & \textbf{0.755 $\pm$ 0.027}
& \textbf{0.935 $\pm$ 0.087} & 0.493 $\pm$ 0.374 
& 0.430 $\pm$ 0.260 & 0.812 $\pm$ 0.132 \\

\bottomrule
\end{tabular}%
}
\end{table*}

\begin{table}[b]
\centering
\caption{Automotive dataset efficiency metrics. Mean $\pm$ Std.\ over 5 seeds. Training and inference times are given for the whole dataset. Best per metric in \textbf{bold}. }
\label{tab:prop_runtime_recall}
\resizebox{\linewidth}{!}{%
\begin{tabular}{@{}lccc@{}}
\toprule
\textbf{Model} & \textbf{Recall@1\%} & \textbf{Train Time (s)} & \textbf{Inference Time (s)} \\
\midrule
VASP~\cite{he2021vasp}               & $0.012 \pm 0.000$ & $42.6 \pm 0.1$     & $52.5 \pm 0.5$ \\
Anomaly Transformer~\cite{xu2022anomalytransformertimeseries} & $0.023 \pm 0.011$ & $69.7 \pm 10.8$    & $56.4 \pm 1.5$ \\
Isolation Forest~\cite{liu2008isolation}   & $0.024 \pm 0.010$ & \textbf{0.7 $\pm$ 0.1} & \textbf{52.1 $\pm$ 4.0} \\
W-VAE~\cite{pereira2019wvae}         & $0.049 \pm 0.005$ & $79.5 \pm 0.7$     & $58.7 \pm 0.5$ \\
VS-VAE~\cite{pereira2018vsvae}       & $0.050 \pm 0.009$ & $102.5 \pm 2.5$    & $62.6 \pm 0.8$ \\
OmniAnomaly~\cite{su2019omni}        & $0.094 \pm 0.003$ & $605.5 \pm 0.8$    & $135.2 \pm 0.5$ \\
SIS-VAE~\cite{li2021sisvae}          & $0.085 \pm 0.004$ & $1081.8 \pm 3.3$   & $130.8 \pm 0.5$ \\
MA-VAE~\cite{correia2023MAVAE}       & $0.102 \pm 0.004$ & $320.3 \pm 13.8$   & $119.6 \pm 4.2$ \\
GDN~\cite{deng2021gdn}               & $0.109 \pm 0.001$ & $456.0 \pm 4.9$    & $130.4 \pm 1.1$ \\
TFT-Residual~\cite{lim2021tft}       & \textbf{0.110 $\pm$ 0.000} & $169.1 \pm 4.4$   & $79.3 \pm 0.9$ \\
\midrule
STREAM-VAE (ours)  & \textbf{0.110 $\pm$ 0.000} & $262.3 \pm 18.0$ & $97.9 \pm 5.4$ \\
\bottomrule
\end{tabular}%
}
\end{table}

Our experimental results are given in Table~\ref{tab:combined_oracle_auc}. 
On the automotive dataset, STREAM-VAE achieves the highest {Oracle PA-F1} (0.857), 
{PA-F1} (0.794), {AUC-PR} (0.532), and a competitive {AUC-ROC} (0.755), 
indicating both strong score ordering and stable detection at a fixed threshold. On SMD, where 
quasi-periodic structure benefits association-based models, {Anomaly Transformer} 
obtains the best {AUC-PR} (0.462) and {AUC-ROC} (0.881), while STREAM-VAE 
remains competitive in the threshold-free metrics and achieves the highest 
{Oracle PA-F1} (0.935), suggesting that its scores remain well ordered even when a 
single global threshold is not perfectly tuned. \looseness=-1

On the automotive data, the performance gain of STREAM-VAE over VAE baselines can be traced to three effects. First, separating fast spikes and slow drifts in the encoder prevents sharp events from being 
oversmoothed by slow context. Second, per-feature mixture of experts in the decoder allows 
benign operating mode changes to be absorbed by expert reweighting instead of by inflating variance. 
Third, the soft-thresholded event residual explains sparse transients additively in the mean, 
which sharpens point-adjusted F1 by reducing fragmented detections and keeps nominal tails 
tight enough for stable POT calibration.

The behavior of the baselines is consistent with their design goals. TFT-Residual performs well 
when anomalies are departures from locally predictable trends, but it is sensitive to unmodeled 
operating mode switches. GDN leverages inter-sensor structure and is strong on correlation anomalies, 
yet its performance can degrade when couplings change with context. SIS-VAE stabilizes 
reconstructions and helps on drifts but tends to attenuate sharp spikes. MA-VAE adds 
long-range context but does not explicitly separate time scales, so variance can grow to 
cover fast deviations. OmniAnomaly’s expressive posterior improves ranking but may handle 
heterogeneity mainly through scale changes, which complicates operation at a fixed threshold. 
VS-VAE and W-VAE capture global structure yet lack explicit routing of operating modes or transients. 
VASP’s sparsity promotes salient spikes but can under-represent slow trends. Anomaly 
Transformer excels on SMD, where periodic anchors repeat, but underperforms on the automotive 
data where associations depend strongly on driving context. Isolation Forest remains competitive at high recall for gross outliers but lacks multivariate temporal modeling, which limits its performance on structured automotive signals. Our ablation study in the following section confirms this interpretation: performance degrades systematically when individual architectural components are removed, indicating that the improvements of STREAM-VAE indeed come from its architectural design.

Finally,  we evaluate computational efficiency on the automotive dataset (Table~\ref{tab:prop_runtime_recall}) for an in-vehicle application. All runtimes were 
measured on a single Apple MacBook Pro M3 Max and refer to processing the full 
dataset. For context, STREAM-VAE processes about 408 windows\,/\,s in Python 
(2.45\,ms per window, $T{=}100$), which corresponds to real-time operation up to 
roughly 400\,Hz. This is well above the 10–100\,Hz update rates of typical automotive 
signals.

To reflect practical on-board in-vehicle requirements, the table reports recall at a fixed 
false-positive rate of 1\%. This metric is not a replacement for the 
threshold-free results shown earlier in Table \ref{tab:combined_oracle_auc}, but an additional operating-point analysis 
chosen because in-vehicle monitors must trigger only rarely while still detecting 
meaningful anomalies. At this operating point, STREAM-VAE 
matches the top recall of TFT-Residual. Its training and inference times fall in 
the mid-range of learned detectors: not as heavy as SIS-VAE, OmniAnomaly, or GDN, 
but naturally slower than very lightweight methods such as Isolation Forest or 
VASP. Anomaly Transformer is efficient but performs poorly at low false-positive 
rates on this dataset.

Overall, STREAM-VAE combines state-of-the-art detection performance with 
computational cost comparable to other modern deep anomaly detectors. Its runtime 
is sufficiently low for real-time in-vehicle monitoring while remaining well-suited 
for large-scale offline fleet analysis.

\begin{table*}[!t]
\centering
\caption{Ablations on the automotive dataset. Mean $\pm$ Std.\ over 5 seeds. Best per metric in \textbf{bold}.}
\label{tab:prop_ablation_oracle_auc}
\begin{tabular}{@{}lcccc@{}}
\toprule
\textbf{Variant} & \multicolumn{1}{c}{\textbf{Oracle PA-F1}} & \multicolumn{1}{c}{\textbf{PA-F1}} & \multicolumn{1}{c}{\textbf{AUC-PR}} & \multicolumn{1}{c}{\textbf{AUC-ROC}} \\
\midrule

STREAM-VAE (Full)              & \textbf{0.792 $\pm$ 0.028} & \textbf{0.631 $\pm$ 0.018} & \textbf{0.309 $\pm$ 0.024} & 0.664 $\pm$ 0.011 \\
w/o MoE Decoder              & 0.785 $\pm$ 0.037          & 0.619 $\pm$ 0.014          & 0.275 $\pm$ 0.034          & 0.658 $\pm$ 0.010 \\
Small Latent (32D)           & 0.784 $\pm$ 0.019          & 0.623 $\pm$ 0.019          & 0.304 $\pm$ 0.024          & 0.665 $\pm$ 0.011 \\
Small MoE (k=4)              & 0.783 $\pm$ 0.016          & 0.617 $\pm$ 0.008          & 0.301 $\pm$ 0.024          & 0.664 $\pm$ 0.012 \\
w/o Input Injection          & 0.781 $\pm$ 0.039          & 0.612 $\pm$ 0.022          & 0.303 $\pm$ 0.036          & 0.661 $\pm$ 0.008 \\
Simple FFN ($\times$1)       & 0.781 $\pm$ 0.021          & 0.604 $\pm$ 0.018          & 0.300 $\pm$ 0.029          & 0.654 $\pm$ 0.006 \\
Concat Merge (No Gate)       & 0.780 $\pm$ 0.034          & 0.614 $\pm$ 0.020          & 0.280 $\pm$ 0.023          & 0.656 $\pm$ 0.006 \\
Drift-Only Attention         & 0.770 $\pm$ 0.034          & 0.611 $\pm$ 0.022          & 0.297 $\pm$ 0.023          & 0.656 $\pm$ 0.015 \\
w/o Event Residuals          & 0.774 $\pm$ 0.032          & 0.614 $\pm$ 0.028          & 0.303 $\pm$ 0.021          & \textbf{0.666 $\pm$ 0.014} \\
Spike-Only Attention         & 0.774 $\pm$ 0.016          & 0.622 $\pm$ 0.014          & 0.254 $\pm$ 0.012          & 0.642 $\pm$ 0.008 \\
w/o Input Injection + MoE    & 0.770 $\pm$ 0.021          & 0.623 $\pm$ 0.016          & 0.289 $\pm$ 0.035          & 0.661 $\pm$ 0.007 \\
w/o Attention                & 0.753 $\pm$ 0.018          & 0.601 $\pm$ 0.013          & 0.279 $\pm$ 0.022          & 0.658 $\pm$ 0.007 \\

\bottomrule
\end{tabular}%
\end{table*}

\begin{figure*}[!tb]
  \centering
  \includegraphics[width=0.73\textwidth]{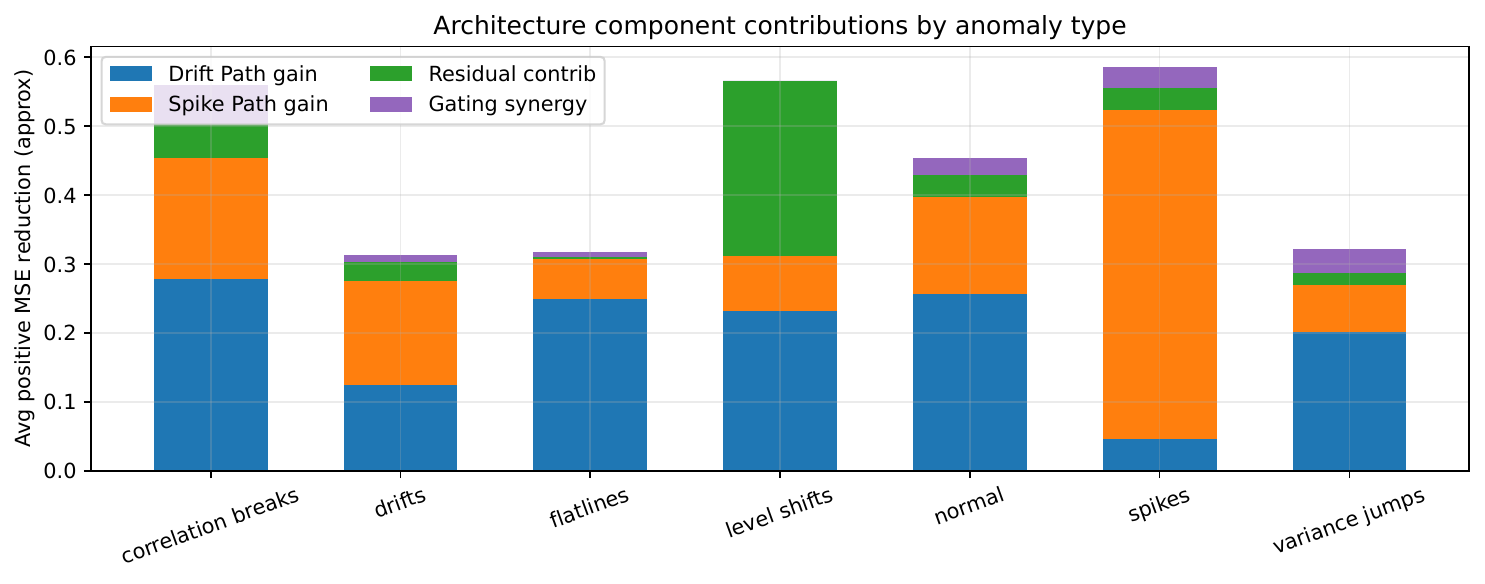}
  \caption{Component-wise contribution to reconstruction error by anomaly type.}
  \label{fig:component_contributions}
\end{figure*}

\section{Ablation Studies}
\label{sec:ablation}

Table~\ref{tab:prop_ablation_oracle_auc} summarizes the effect of removing or
simplifying individual components of STREAM-VAE. The full model leads in both
threshold-free separation and Oracle Point-Adjusted detection. Removing the
event residual lowers both Point-Adjusted F1 and AUC-PR while producing a small
increase in AUC-ROC. This indicates that the residual improves precision-oriented
ranking near operating thresholds while slightly smoothing the extremes of the
score distribution that influence ROC. Disabling the per-feature MoE increases
false alarms during benign operating mode shifts, supporting its role in
explaining away mode changes through expert routing. Collapsing the dual-path
encoder to a single path or removing attention reduces score ordering quality
across thresholds even if a single operating point can look acceptable; explicit
separation of slow drift and fast spike information prevents variance inflation
and drift-spike blurring. Ungated merges, either for injecting the residual or
for combining the drift and spike pathways, tend to overfire and destabilize
calibration, while the gated input injection keeps latents informative so that
attention and MoE can operate effectively. Reduced-capacity variants, such as those with a smaller latent dimension,
fewer experts, or a shallower decoder FFN, track the full model closely,
which indicates that the gains come from the architectural design rather
than parameter count.

To relate these behaviors to how the architecture reacts to specific anomaly
types, we quantify component-wise contributions using a controlled decoder
analysis that matches the metric in Fig.~\ref{fig:component_contributions}. The
drift path and spike path correspond to the attention patterns shown in
Fig.~\ref{fig:streamvae}, where the drift path exhibits stable global anchors
and the spike path displays localized, high-frequency structure. For each window
we encode once and freeze the latent sequence, then reconstruct the window with
different component subsets, such as the drift path only, the spike path only,
or no residual. We compare the reconstruction error (mean squared error, MSE) of
these restricted variants to that of the full model and record only positive
increases in MSE. This measures how much each component helps the model reduce
reconstruction error when present. Fig.~\ref{fig:component_contributions}
reports the average effects across anomaly categories. Spikes are mainly handled
by the spike path, which matches the
short-lag, high-frequency coupling seen in the spike-path attention maps.
Drifts benefit from all components: the spike path
and drift path provide most of the gain, with only a small residual
contribution, consistent with the anchor-based attention patterns in
Fig.~\ref{fig:streamvae}. Level shifts are
led by the residual and the drift path. Variance jumps
are also drift path-led with moderate support from the spike path. Flatlines rely
mainly on the drift path with a moderate spike path contribution. Correlation breaks show
strong, shared improvements from both paths, with the drift path slightly
ahead. On normal windows the contributions are small and
distributed, indicating that the mixture cleans reconstructions without
overusing the residual.

\section{Conclusion}
In this paper, we presented STREAM-VAE, a time-scale-aware variational
autoencoder for anomaly detection in vehicle telemetry. The model centers on a
dual-path encoder that separates slow drift from fast spike dynamics, supported
by a lightweight decoder design that models transient deviations without
confusing them with normal operating behavior. Experiments on automotive data
and a public benchmark show that this separation of time scales improves
robustness while keeping computation suitable for both in-vehicle deployment and
backend fleet analysis. STREAM-VAE therefore provides a compact and practical
basis for reliable telemetry anomaly detection in intelligent vehicles. A
current limitation is that the method relies on per-entity calibration that may shift across vehicles
and environments. This suggests future work on more transferable calibration.

\bibliographystyle{IEEEtran}
\bibliography{references}

@inproceedings{kingma2013vae,
  author       = {Diederik P. Kingma and
                  Max Welling},
  editor       = {Yoshua Bengio and
                  Yann LeCun},
  title        = {Auto-Encoding Variational {Bayes}},
  booktitle    = {2nd International Conference on Learning Representations, {ICLR} 2014,
                  Banff, AB, Canada, April 14-16, 2014, Conference Track Proceedings},
  year         = {2014},
  bibsource    = {dblp computer science bibliography, https://dblp.org}
}

@INPROCEEDINGS{GRAVES2005602,
  author={Graves, A. and Schmidhuber, J.},
  booktitle={Proceedings. 2005 IEEE International Joint Conference on Neural Networks, 2005.}, 
  title={Framewise phoneme classification with bidirectional {LSTM} networks}, 
  year={2005},
  volume={4},
  number={},
  pages={2047-2052 vol. 4},
  doi={10.1109/IJCNN.2005.1556215}}

@inproceedings{su2019omni,
author = {Su, Ya and Zhao, Youjian and Niu, Chenhao and Liu, Rong and Sun, Wei and Pei, Dan},
title = {Robust Anomaly Detection for Multivariate Time Series through Stochastic Recurrent Neural Network},
year = {2019},
isbn = {9781450362016},
publisher = {Association for Computing Machinery},
address = {New York, NY, USA},
doi = {10.1145/3292500.3330672},
booktitle = {Proceedings of the 25th ACM SIGKDD International Conference on Knowledge Discovery \& Data Mining},
pages = {2828–2837},
numpages = {10},
location = {Anchorage, AK, USA},
series = {KDD '19}
}

@ARTICLE{li2021sisvae,
  author={Li, Longyuan and Yan, Junchi and Wang, Haiyang and Jin, Yaohui},
  journal={IEEE Transactions on Neural Networks and Learning Systems}, 
  title={Anomaly Detection of Time Series With Smoothness-Inducing Sequential Variational Auto-Encoder}, 
  year={2021},
  volume={32},
  number={3},
  pages={1177-1191},
  doi={10.1109/TNNLS.2020.2980749}}

@article{pereira2019wvae,
author = {Pereira, Jo{{\~a}o} and Silveira, Margarida},
title = {Unsupervised representation learning and anomaly detection in {ECG} sequences},
year = {2019},
issue_date = {2019},
publisher = {Inderscience Publishers},
address = {Geneva 15, CHE},
volume = {22},
number = {4},
issn = {1748-5673},
doi = {10.1504/ijdmb.2019.101395},
journal = {Int. J. Data Min. Bioinformatics},
month = jan,
pages = {389–407},
numpages = {18}
}

@article{he2021vasp,
title = {{VASP}: An autoencoder-based approach for multivariate anomaly detection and robust time series prediction with application in motorsport},
journal = {Engineering Applications of Artificial Intelligence},
volume = {104},
pages = {104354},
year = {2021},
issn = {0952-1976},
doi = {https://doi.org/10.1016/j.engappai.2021.104354},
author = {Julian {von Schleinitz} and Michael Graf and Wolfgang Trutschnig and Andreas Schröder}
}

@inproceedings{bahuleyan2018,
    title = "Variational Attention for Sequence-to-Sequence Models",
    author = "Bahuleyan, Hareesh  and
      Mou, Lili  and
      Vechtomova, Olga  and
      Poupart, Pascal",
    editor = "Bender, Emily M.  and
      Derczynski, Leon  and
      Isabelle, Pierre",
    booktitle = "Proceedings of the 27th International Conference on Computational Linguistics",
    month = aug,
    year = "2018",
    address = "Santa Fe, New Mexico, USA",
    publisher = "Association for Computational Linguistics",
    pages = "1672--1682"
}

@INPROCEEDINGS{pereira2018vsvae,
  author={Pereira, Jo{{\~a}o} and Silveira, Margarida},
  booktitle={2018 17th IEEE International Conference on Machine Learning and Applications (ICMLA)}, 
  title={Unsupervised Anomaly Detection in Energy Time Series Data Using Variational Recurrent Autoencoders with Attention}, 
  year={2018},
  volume={},
  number={},
  pages={1275-1282},
  doi={10.1109/ICMLA.2018.00207}}

@article{ainslie2023gqa,
  title={{GQA}: Training Generalized Multi-Query Transformer Models from Multi-Head Checkpoints},
  author={Joshua Ainslie and James Lee-Thorp and Michiel de Jong and Yury Zemlyanskiy and Federico Lebr{{\'o}n} and Sumit K. Sanghai},
  journal={ArXiv},
  year={2023},
  volume={abs/2305.13245},
}

@inproceedings{rezende2015flows,
author = {Rezende, Danilo Jimenez and Mohamed, Shakir},
title = {Variational inference with normalizing flows},
year = {2015},
publisher = {JMLR.org},
booktitle = {Proceedings of the 32nd International Conference on International Conference on Machine Learning - Volume 37},
pages = {1530–1538},
numpages = {9},
location = {Lille, France},
series = {ICML'15}
}

@conference{correia2023mavae,
author={Lucas Correia and Jan{-}Christoph Goos and Philipp Klein and Thomas Bäck and Anna Kononova},
title={{MA-VAE}: Multi-Head Attention-Based Variational Autoencoder Approach for Anomaly Detection in Multivariate Time-Series Applied to Automotive Endurance Powertrain Testing},
booktitle={Proceedings of the 15th International Joint Conference on Computational Intelligence - NCTA},
year={2023},
pages={407-418},
publisher={SciTePress},
organization={INSTICC},
doi={10.5220/0012163100003595},
isbn={978-989-758-674-3},
issn={2184-3236},
}

@book{coles2001extreme,
  title = {An Introduction to Statistical Modeling of Extreme Values},
  ISBN = {9781447136750},
  ISSN = {2197-568X},
  DOI = {10.1007/978-1-4471-3675-0},
  journal = {Springer Series in Statistics},
  publisher = {Springer London},
  author = {Coles,  Stuart},
  year = {2001}
}

@inproceedings{shao2020controlvae,
author = {Shao, Huajie and Yao, Shuochao and Sun, Dachun and Zhang, Aston and Liu, Shengzhong and Liu, Dongxin and Wang, Jun and Abdelzaher, Tarek},
title = {ControlVAE: controllable variational autoencoder},
year = {2020},
publisher = {JMLR.org},
booktitle = {Proceedings of the 37th International Conference on Machine Learning},
articleno = {803},
numpages = {10},
series = {ICML'20}
}

@Article{zhao2022dualstage,
AUTHOR = {Zhao, Yun and Zhang, Xiuguo and Shang, Zijing and Cao, Zhiying},
TITLE = {{DA-LSTM-VAE}: Dual-Stage Attention-Based {LSTM}-{VAE} for {KPI} Anomaly Detection},
JOURNAL = {Entropy},
VOLUME = {24},
YEAR = {2022},
NUMBER = {11},
ARTICLE-NUMBER = {1613},
PubMedID = {36359702},
ISSN = {1099-4300},
DOI = {10.3390/e24111613}
}

@inproceedings{deng2021gdn,
  title={Graph neural network-based anomaly detection in multivariate time series},
  author={Deng, Ailin and Hooi, Bryan},
  booktitle={Proceedings of the AAAI Conference on Artificial Intelligence},
  volume={35},
  number={5},
  pages={4027--4035},
  year={2021}
}

@INPROCEEDINGS{liu2008isolation,
  author={Liu, Fei Tony and Ting, Kai Ming and Zhou, Zhi-Hua},
  booktitle={2008 Eighth IEEE International Conference on Data Mining}, 
  title={Isolation Forest}, 
  year={2008},
  volume={},
  number={},
  pages={413-422},
  doi={10.1109/ICDM.2008.17}}

@article{lim2021tft,
title = {Temporal Fusion Transformers for interpretable multi-horizon time series forecasting},
journal = {International Journal of Forecasting},
volume = {37},
number = {4},
pages = {1748-1764},
year = {2021},
issn = {0169-2070},
doi = {https://doi.org/10.1016/j.ijforecast.2021.03.012},
author = {Bryan Lim and Sercan \"O. Ar{\i}k and Nicolas Loeff and Tomas Pfister}
}

@inproceedings{xu2022anomalytransformertimeseries,
title={Anomaly Transformer: Time Series Anomaly Detection with Association Discrepancy},
author={Jiehui Xu and Haixu Wu and Jianmin Wang and Mingsheng Long},
booktitle={International Conference on Learning Representations},
year={2022},
}

@inproceedings{xu2018kpi,
author = {Xu, Haowen and Chen, Wenxiao and Zhao, Nengwen and Li, Zeyan and Bu, Jiahao and Li, Zhihan and Liu, Ying and Zhao, Youjian and Pei, Dan and Feng, Yang and Chen, Jie and Wang, Zhaogang and Qiao, Honglin},
title = {Unsupervised Anomaly Detection via Variational Auto-Encoder for Seasonal {KPIs} in Web Applications},
year = {2018},
isbn = {9781450356398},
publisher = {International World Wide Web Conferences Steering Committee},
address = {Republic and Canton of Geneva, CHE},
doi = {10.1145/3178876.3185996},
booktitle = {Proceedings of the 2018 World Wide Web Conference},
pages = {187–196},
numpages = {10},
location = {Lyon, France},
series = {WWW '18}
}

@article{rousseeuw1993alternatives,
 ISSN = {01621459, 1537274X},
 author = {Peter J. Rousseeuw and Christophe Croux},
 journal = {Journal of the American Statistical Association},
 number = {424},
 pages = {1273--1283},
 publisher = {[American Statistical Association, Taylor & Francis, Ltd.]},
 title = {Alternatives to the Median Absolute Deviation},
 urldate = {2025-11-14},
 volume = {88},
 year = {1993}
}

@ARTICLE{donoho1995denoising,
  author={Donoho, D.L.},
  journal={IEEE Transactions on Information Theory}, 
  title={De-noising by soft-thresholding}, 
  year={1995},
  volume={41},
  number={3},
  pages={613-627},
  doi={10.1109/18.382009}}

@ARTICLE{jacobs1991adaptive,
  author={Jacobs, Robert A. and Jordan, Michael I. and Nowlan, Steven J. and Hinton, Geoffrey E.},
  journal={Neural Computation}, 
  title={Adaptive Mixtures of Local Experts}, 
  year={1991},
  volume={3},
  number={1},
  pages={79-87},
  doi={10.1162/neco.1991.3.1.79}}

@INPROCEEDINGS{jordan1994hmoe,
  author={Jordan, M.I. and Jacobs, R.A.},
  booktitle={Proceedings of 1993 International Conference on Neural Networks (IJCNN-93-Nagoya, Japan)}, 
  title={Hierarchical mixtures of experts and the {EM} algorithm}, 
  year={1993},
  volume={2},
  number={},
  pages={1339-1344 vol.2},
  doi={10.1109/IJCNN.1993.716791}}

@inproceedings{he2016deep,
  author={He, Kaiming and Zhang, Xiangyu and Ren, Shaoqing and Sun, Jian},
  booktitle={2016 IEEE Conference on Computer Vision and Pattern Recognition (CVPR)}, 
  title={Deep Residual Learning for Image Recognition}, 
  year={2016},
  volume={},
  number={},
  pages={770-778},
  doi={10.1109/CVPR.2016.90}}

@inproceedings{siffer2017anomaly,
author = {Siffer, Alban and Fouque, Pierre-Alain and Termier, Alexandre and Largouet, Christine},
title = {Anomaly Detection in Streams with Extreme Value Theory},
year = {2017},
isbn = {9781450348874},
publisher = {Association for Computing Machinery},
address = {New York, NY, USA},
doi = {10.1145/3097983.3098144},
booktitle = {Proceedings of the 23rd ACM SIGKDD International Conference on Knowledge Discovery and Data Mining},
pages = {1067–1075},
numpages = {9},
location = {Halifax, NS, Canada},
series = {KDD '17}
}

@inproceedings{vaswani2023attentionneed,
author = {Vaswani, Ashish and Shazeer, Noam and Parmar, Niki and Uszkoreit, Jakob and Jones, Llion and Gomez, Aidan N. and Kaiser, \L{}ukasz and Polosukhin, Illia},
title = {Attention is all you need},
year = {2017},
booktitle = {Proceedings of the 31st International Conference on Neural Information Processing Systems},
pages = {6000–6010},
numpages = {11},

}

@inproceedings{akiba2019optuna,
  title={{O}ptuna: A Next-Generation Hyperparameter Optimization Framework},
  author={Akiba, Takuya and Sano, Shotaro and Yanase, Toshihiko and Ohta, Takeru and Koyama, Masanori},
  booktitle={The 25th ACM SIGKDD International Conference on Knowledge Discovery \& Data Mining},
  pages={2623--2631},
  year={2019}
}

@inproceedings{kingma2015adam,
  author = {Kingma, Diederik P. and Ba, Jimmy},  
  booktitle = {ICLR},
  title = {Adam: A Method for Stochastic Optimization.},
  year = 2015
}
\newpage
\appendices

\end{document}